\documentclass[letterpaper]{article} 
\usepackage{aaai2026}  
\usepackage{times}  
\usepackage{helvet}  
\usepackage{courier}  
\usepackage[hyphens]{url}  
\usepackage{graphicx} 
\usepackage{natbib}  
\usepackage{caption} 
\usepackage{algorithm}
\usepackage{algorithmic}
\usepackage{graphicx}
\usepackage{amsmath} 
\usepackage{amssymb} 
\usepackage{booktabs}
\usepackage{multirow}
\usepackage{float} 
\usepackage{amsmath}
\usepackage{newfloat}
\usepackage{listings}
\DeclareCaptionStyle{ruled}{labelfont=normalfont,labelsep=colon,strut=off} 
\floatstyle{ruled}
\newfloat{listing}{tb}{lst}{}
\floatname{listing}{Listing}
\title{Modeling Uncertainty Trends for Timely Retrieval in Dynamic RAG}
\author{
    Bo Li\textsuperscript{\rm 1, 3},
    Tian Tian\textsuperscript{\rm 1},
    Zhenghua Xu\textsuperscript{\rm 1}\equalcontrib,
    Hao Cheng\textsuperscript{\rm 2},
    Shikun Zhang\textsuperscript{\rm 3},
    Wei Ye\textsuperscript{\rm 3}\equalcontrib
}
\affiliations{
    \textsuperscript{\rm 1}State Key Laboratory of Intelligent Power Distribution Equipment and System, \\School of Health Sciences and Biomedical Engineering, Hebei University of Technology, China\\
    \textsuperscript{\rm 2}School of Artificial Intelligence, Hebei University of Technology, China \\
    \textsuperscript{\rm 3}National Engineering Research Center for Software Engineering, Peking University, China\\
    \texttt{deepblue.lb@gmail.com, zhenghua.xu@hebut.edu.cn, wye@pku.edu.cn}
}

\usepackage{bibentry}

\begin{document}

\maketitle

\begin{abstract}

Dynamic retrieval-augmented generation (RAG) allows large language models (LLMs) to fetch external knowledge on demand, offering greater adaptability than static RAG. A central challenge in this setting lies in determining the optimal timing for retrieval. Existing methods often trigger retrieval based on low token-level confidence, which may lead to delayed intervention after errors have already propagated. We introduce Entropy-Trend Constraint (ETC), a training-free method that determines optimal retrieval timing by modeling the dynamics of token-level uncertainty. Specifically, ETC utilizes first- and second-order differences of the entropy sequence to detect emerging uncertainty trends, enabling earlier and more precise retrieval. Experiments on six QA benchmarks with three LLM backbones demonstrate that ETC consistently outperforms strong baselines while reducing retrieval frequency. ETC is particularly effective in domain-specific scenarios, exhibiting robust generalization capabilities. Ablation studies and qualitative analyses further confirm that trend-aware uncertainty modeling yields more effective retrieval timing. The method is plug-and-play, model-agnostic, and readily integrable into existing decoding pipelines. Implementation code is included in the supplementary materials.


\end{abstract}

\begin{links}
    \link{Code}{https://github.com/pkuserc/ETC}
\end{links}


\section{Introduction}

Retrieval-Augmented Generation (RAG) has emerged as a powerful paradigm for augmenting large language models (LLMs) with external knowledge, effectively addressing limitations such as outdated training data and narrow domain coverage~\cite{gao2023retrieval,kandpal2023large,mousavi2024your,xiong2024search}. By incorporating retrieved documents during generation, RAG systems significantly improve factual accuracy and enhance generalization across domains~\cite{xu2024unsupervisedir,fang2024enhancingnr}. While early RAG systems typically performed a single retrieval at the start of generation~\cite{wang2024biorag,Shi2023REPLUGRB,Wang2023Query2docQE,Yu2023ImprovingLM}, recent developments have introduced dynamic RAG, where retrieval is triggered conditionally during decoding to balance informativeness and efficiency~\cite{jiang2023active,su2024dragin}.

\begin{figure}[t]
    \centering
    \includegraphics[width=1.0\linewidth]{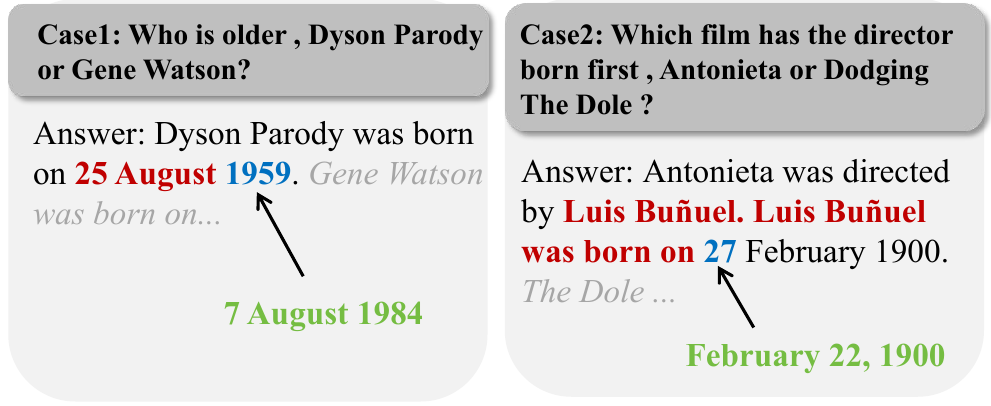}
    \caption{The delayed retrieval issue exists in current dynamic RAG method, where blue tokens represent DRAGIN's retrieval timing, and red tokens highlight incorrectly generated tokens caused by delayed retrieval.}
    \label{pic.intro}
\end{figure}

A central challenge in dynamic RAG is deciding when retrieval should occur. Previous approaches mainly rely on token-level uncertainty heuristics. For instance, some methods~\cite{borgeaud2022improving,trivedi2022interleaving,ram2023context} perform retrieval after a fixed number of tokens or sentences, while others trigger retrieval when the confidence of a newly generated token drops below a predefined threshold~\cite{jiang2023active,su2024dragin}. Although intuitive, such reactive mechanisms often suffer from delayed retrieval, i.e., retrieving only after the model has already deviated from the correct generation path. As illustrated in Figure~\ref{pic.intro}, retrieval triggered too late may fail to prevent factual errors, whereas overly early or frequent retrieval increases latency and redundancy~\cite{ni2024whendl,ren2023investigating,chen2024benchmarking,maekawa2024retrieval}.

We argue that retrieval timing should be guided not by isolated token-level confidence values, but by tracking the overall trend of uncertainty throughout the generation process. Foundational studies on LLMs show that entropy-based uncertainty measures are more robust and informative than pointwise confidence for detecting hallucinations or unreliable generation. For instance, recent work has employed entropy as a signal for token-level factuality assessment, semantic instability, and fine-grained uncertainty estimation~\cite{Fadeeva2024FactCheckingTO,Farquhar2024DetectingHI,Nikitin2024KernelLE}. These findings suggest that tracking how uncertainty evolves over time yields more reliable signals than reacting to isolated confidence drops at individual tokens. This observation points to a promising direction: modeling the dynamics of token-level uncertainty to improve retrieval decisions during generation.

Building on this insight, we propose \textbf{E}ntropy-\textbf{T}rend \textbf{C}onstraint (\textbf{ETC}), a novel training-free method that models the dynamics of uncertainty throughout generation rather than relying on individual token-level confidence. Specifically, ETC analyzes the first- and second-order differences of the token-level entropy sequence to detect emerging low-confidence patterns before they become critical. These differential operations are classical tools for discrete sequence analysis~\cite{jordan1965calculus,levy1992finite,ames2014numerical}. In particular, the second-order difference provides a sensitive signal for detecting rapid shifts in entropy, indicating that the model may be entering an unstable prediction phase. To enhance robustness, we further introduce a dynamic smoothing mechanism to reduce the impact of entropy outliers and stabilize retrieval decisions. By leveraging confidence trends, ETC enables timely retrieval, injecting external knowledge at more optimal positions while reducing retrieval frequency. Unlike existing methods that rely on heuristic rules or costly training procedures, ETC is plug-and-play, model-agnostic, and easily integrable into any autoregressive decoding pipeline.

We evaluate ETC on six diverse benchmarks spanning multi-hop reasoning, commonsense QA, reading comprehension, and biomedical QA. Across three LLM backbones, ETC consistently outperforms strong baselines while requiring fewer retrieval operations and achieving significantly lower rates of delayed and redundant retrieval. Furthermore, qualitative evaluations using GPT-4o and extensive ablation studies validate the precision and efficiency of ETC’s retrieval timing strategy.

\begin{itemize}
    \item We identify and systematically analyze the delayed retrieval problem in dynamic RAG, revealing fundamental limitations in confidence-based triggering strategies.
    \item We propose \textbf{E}ntropy-\textbf{T}rend \textbf{C}onstraint (ETC), a novel training-free retrieval strategy that leverages uncertainty dynamics for timely and efficient knowledge injection.
    \item Experiments on six diverse benchmarks with three LLM backbones demonstrate that ETC consistently improves performance while reducing retrieval frequency. We further present comprehensive analyses and case studies to support these findings.
\end{itemize}

\section{Preliminary and Delayed Retrieval}
In this section, we briefly introduce the fundamentals of RAG and the issue of delayed retrieval.

\subsection{Preliminary}

In a standard RAG system, for a given query \textit{q}, a retriever \textit{r} retrieves a set of relevant documents $C = \{c_1, c_2, .., c_n\}$ from a large corpus \textit{D}. During inference, the query \textit{q} and retrieved contexts \textit{C} are combined through a prompt \textit{p} to form a new input for a given LLM, which then generates the retrieval-augmented output \textit{y}. Typically, retrieval is performed once at the beginning of the generation process~\cite{melz2023enhancing,wang2024biorag,li2024enhancing}. The mathematical expression of this process is shown as follows:

\begin{equation}
    y = LLM(q, C, p).
\end{equation}

However, studies have shown that retrieving information solely at the beginning of generation may not always be optimal, as irrelevant or redundant context can introduce noise and hinder model performance. An alternative approach is Dynamic RAG, which performs retrieval only when needed. Recent methods typically trigger retrieval when the model generates a token with very low confidence. Let time \textit{t} denote the point at which retrieve is triggered, indicating that external knowledge is required after generating the \textit{t}-th token. The prompt at the step $t$ is denoted as $p_{t}$. The dynamic RAG model can then be formally expressed as:

\begin{equation}
    \hat{y} = LLM(q, C_{t}, p_{t}, y_{<t}),
\end{equation}
where $y_{<t}$ represents the sequence of tokens generated before time \textit{t}, and $C_{t}$ is the external knowledge retrieved after the \textit{t}-th token has been generated.

\subsection{Delayed Retrieval}

While using the confidence of a single token to trigger retrieval is a straightforward approach, it does not always lead to optimal retrieval timing. Through both qualitative and quantitative analyses, we identify a delayed retrieval problem that arises when retrieval timing is determined primarily by token-level confidence. Specifically, we examine cases from the 2WikiMultihopQA~\cite{ho2020constructing} dataset using the DRAGIN framework~\cite{su2024dragin}. As shown in Figure~\ref{pic.intro}, retrieval is triggered when a generated token falls below DRAGIN’s predefined confidence threshold. These examples clearly illustrate that by the time retrieval is triggered, the generation has already deviated from the correct path. Consequently, this approach often results in multiple low-confidence tokens being generated before retrieval, suggesting that confidence-based intervention may not be the most effective solution.

In addition, we manually annotate 100 instances to compute the proportion of delayed retrieval, and find that around 33\% of them exhibit this issue (see \textbf{Section}~\ref{sec.delay} for more details). These findings highlight the need for approaches that are more sensitive to token-level confidence trends during generation.


\section{Proposed Method}

Building on the above analysis, we propose \textbf{E}ntropy-\textbf{T}rend \textbf{C}onstraint (\textbf{ETC}), a novel dynamic RAG method that mitigates delayed retrieval by considering the confidence trend of the generated sequence. Specifically, ETC computes confidence trends based on the entropy sequence and triggers retrieval when confidence changes sharply. In addition, we introduce a dynamic smoothing strategy that reduces unnecessary retrievals while maintaining model performance.


\subsection{Entropy-Trend Constraint}
Given an input $c$ and a prompt $p$, a LLM generates an output token sequence, denoted as $T = \{t_1, t_2,..., t_n\}$, where the length of $T$ is $n$. For each generated token $t_i$, we compute its prediction distribution $p_i(v)$ over the vocabulary $\mathcal{V}$, and use entropy as a measure of its prediction uncertainty, defined as follows:

\begin{equation}
    \mathcal{H}_i = - \sum_{v \in \mathcal{V}} p_i(v) \log p_i(v).
\end{equation}

Entropy is widely used in natural language processing tasks to quantify the uncertainty of a given probability distribution. A lower entropy value indicates higher confidence in the LLM’s prediction. For the generated output $T$, we define its entropy sequence $\mathcal{H}$ as follows:

\begin{equation}
    \mathcal{H} = \{\mathcal{H}_1, \mathcal{H}_2, ..., \mathcal{H}_n\}.
\end{equation}

The entropy sequence $\mathcal{H}$ reflects the confidence associated with each generated token. However, static entropy values alone do not capture dynamic fluctuations during the generation process. Therefore, we utilize both first and second differences of the entropy sequence. The first difference is defined as the difference between consecutive terms. This operation measures the change between successive elements in the entropy sequence and is useful for identifying trends and linearity. For the entropy sequence $\mathcal{H}$ of the generated tokens, the first difference is computed as bellow:

\begin{equation}
    \Delta \mathcal{H} =  \{\Delta \mathcal{H}_1, \Delta 
 \mathcal{H}_2, ..., \Delta  \mathcal{H}_{n-1}\},
\end{equation}

\begin{equation}
    \Delta \mathcal{H}_i = \mathcal{H}_{i+1} - \mathcal{H}_{i}.
\end{equation}

For example, $\Delta \mathcal{H}_1 = \mathcal{H}_{2} - \mathcal{H}_{1}$, and $\Delta \mathcal{H}$ has $n-1$ items. While the first difference reflects the variation between adjacent time points, it does not reveal how rapidly these changes occur. Monitoring the rate of change in these trends is crucial for timely retrieval, as it helps detect early signs of instability in the model’s confidence. Thus we further compute the second difference of the entropy sequence $\mathcal{H}$ to capture the rate of confidence change:

\begin{equation}
    \Delta^2 \mathcal{H} = \Delta(\Delta \mathcal{H}) = \{\Delta^2 \mathcal{H}_1, \Delta^2 
 \mathcal{H}_2, ..., \Delta^2  \mathcal{H}_{n-2}\},
\end{equation}

\begin{equation}
    \Delta^2 \mathcal{H}_i = \Delta \mathcal{H}_{i+1} - \Delta \mathcal{H}_{i}.
\end{equation}

Alternatively, we can also obtain the $\Delta^2 \mathcal{H}$ based on $\mathcal{H}$ directly, where:

\begin{equation}
    \Delta^2 \mathcal{H}_i = \mathcal{H}_{i+2} - 2\mathcal{H}_{i+1} + \mathcal{H}_{i}.
\end{equation}

The second difference highlights rapid shifts in confidence and thus serves as a more sensitive indicator for triggering retrieval.

\subsection{Dynamic Smoothing Method}

While the second difference improves retrieval timing, it can still be influenced by outlier entropy values from specific tokens, potentially leading to redundant retrievals. To address this issue, we propose a dynamic smoothing method that reduces the impact of outliers by assigning them lower weights during aggregation. Specifically, dynamic smoothing computes the $\Delta^2 \mathcal{\widehat{H}}_t$ using weighted average of $\Delta^2 \mathcal{H}_{t}$ and $\Delta^2 \mathcal{H}_{t-1}$, the weight $w_t$ of $\Delta^2 \mathcal{H}_{t}$ is obtained by the following equation:

\begin{equation}
    w_t = \frac{|\Delta^{2} \mathcal{H}_{t-1} - E_t|}{
|\Delta^{2} \mathcal{H}_t-E_t| + |\Delta^{2} \mathcal{H}_{t-1}-E_t|},
\end{equation}

\begin{equation}
    E_t = \mathbb{E}[\Delta^2 \mathcal{H}_1, \Delta^2 
 \mathcal{H}_2, ..., \Delta^2  \mathcal{H}_{t}].
\end{equation}

Here, $E_t$ denotes the mathematical expectation of the entropy sequence's second difference $\{\Delta^{2}\mathcal{H}_1, \Delta^{2}\mathcal{H}_2, ..., \Delta^{2}\mathcal{H}_t\}$. A higher $\Delta^2 \mathcal{H}_t$ results in a relatively lower $w_t$, thereby reducing the impact of outlier entropy and producing a smoothed $\Delta^2 \mathcal{\widehat{H}}_t$ at the current timing $t$. The smoothed $\Delta^2 \mathcal{\widehat{H}}_t$ is computed as follows:

\begin{equation}
    \Delta^2 \mathcal{\widehat{H}}_t = w_t\Delta^2 \mathcal{H}_t + w_{t-1}\Delta^2 \mathcal{H}_{t-1},
\end{equation}

By mitigating the influence of the outlier entropy, the smoothed second difference $\Delta^2 \mathcal{\widehat{H}}_t$ reduces unnecessary retrievals. We perform a retrieval operation at time $t$ if $|\Delta^2 \mathcal{\widehat{H}}_t| \geq \alpha$, where $\alpha$ is a predefined threshold and will be tuned on the validation set.

\subsection{Query Construction and Continue Generation}

Once ETC determines the optimal retrieval timing, the next challenge is to construct an effective query based on the original input and the generated text so far. We adopt the query construction method from DRAGIN, which leverages the self-attention mechanism of Transformer-based LLMs. This approach ranks tokens by their attention scores and selects the \textit{top-n} tokens to form the query\footnote{Due to space limitations, please refer to~\citet{su2024dragin} for further details.}. After constructing the query at timestep \( t \), ETC retrieves relevant information \( C_t = \{C_t^1, C_t^2, \dots\} \) from the external corpus, where \( C_t^i \) denotes the \( i \)-th retrieved document.

To continue generation after retrieval, we combine the original query \( q \), the previously generated text \( y_{<t} \), the retrieved information \( C_t \), and the prompt \( p_t \) to guide the LLM in generating the subsequent output. The generated token \( y_t \) serves as the prefix for subsequent generation, ensuring that the output remains coherent and consistent with the prior sequence \( y_{<t} \).

\section{Experimental Setup}
\subsection{Dataset and Evaluation Metic}

\begin{table}[h]
\centering
\begin{tabular}{@{}c|c|c@{}}
\toprule[1.5pt]
Dataset         & Domain      &  Metric \\ \midrule[1.5pt]
2WikiMultihopQA & Multi-hop   & EM, F1            \\
HotpotQA        & Multi-hop   & EM, F1            \\
StrategyQA      & Commonsense & Accuracy          \\
IIRC            & Reading     & EM, F1            \\\hline
BioASQ          & Biomedical  & Accuracy          \\ 
PubMedQA        & Biomedical     & Accuracy          \\
\bottomrule[1.5pt]
\end{tabular}
\caption{The statistics of the datasets used in this study, and we adopt the same metrics as previous works.}
\label{table.data}
\end{table}

To comprehensively evaluate the effectiveness of ETC across diverse scenarios, we conduct experiments on six representative datasets: 2WikiMultihopQA~\cite{ho2020constructing}, HotpotQA~\cite{yang2018hotpotqa}, StrategyQA~\cite{geva2021did}, IIRC~\cite{ferguson2020iirc}, BioASQ~\cite{tsatsaronis2015overview}, and PubMedQA~\cite{Jin2019PubMedQAAD}. The domains and evaluation metrics for each dataset are summarized in Table~\ref{table.data}. Specifically, the first four datasets are used for general-purpose multi-hop and commonsense QA evaluation, while the last two assess ETC's effectiveness in settings that require biomedical domain knowledge under limited-resource conditions.

\begin{table*}[!h]
\centering
\begin{tabular}{c|c|cc|cc|c|cc|c}
\toprule[1.5pt]
                                     &                          & \multicolumn{2}{c|}{\textbf{2WikiMultihopQA}} & \multicolumn{2}{c|}{\textbf{HotpotQA}} & \textbf{StrategyQA} & \multicolumn{2}{c|}{\textbf{IIRC}}&  \\ \hline
\textbf{LLM}                         & \textbf{RAG Method}      & \textbf{EM}           & \textbf{F1}          & \textbf{EM}       & \textbf{F1}       & \textbf{Accuracy}   & \textbf{EM}     & \textbf{F1} & \textbf{Avg.Score}     \\\midrule[1.5pt]
\multirow{7}{*}{\textbf{Llama2-7b}}  & \textbf{w/o RAG}         & 0.146                 & 0.223                & 0.184             & 0.275             & \textbf{0.659}      & 0.139           & 0.173   & 0.257       \\
                                     & \textbf{Single RAG}      & 0.169                 & 0.255                & 0.164             & 0.250             & 0.645               & 0.187           & 0.226     &0.271       \\
                                     & \textbf{In-Context RALM} & 0.112                 & 0.192                & 0.146             & 0.211             & 0.635               & 0.172           & 0.202    & 0.239        \\
                                     & \textbf{IRCoT}           & 0.189                 & 0.265                & 0.214             & 0.304             & 0.630               & 0.178           & 0.216     & 0.285       \\
                                     & \textbf{FLARE}           & 0.143                 & 0.213                & 0.149             & 0.221             & 0.627               & 0.136           & 0.164      & 0.236      \\
                                     & \textbf{DRAGIN}          & 0.220                 & 0.293                & 0.232             & 0.334             & 0.641               & 0.192           & 0.234     & \underline{0.307}       \\
                                     & \textbf{ETC(Ours)}       & \textbf{0.271}        & \textbf{0.360}       & \textbf{0.288}    & \textbf{0.401}    & 0.650               & \textbf{0.199}  & \textbf{0.240} & \textbf{0.344\textsubscript{\scriptsize{+12.1\%}}}  \\\hline
\multirow{7}{*}{\textbf{Llama3-8b}} & \textbf{w/o RAG}         & 0.174                 & 0.258                & 0.281             & 0.379             & 0.667               & 0.187           & 0.223         &0.310   \\
                                     & \textbf{Single RAG}      & 0.241                 & 0.349                & \textbf{0.339}             & 0.454             & 0.651               & 0.275           & 0.316         & \underline{0.375}   \\
                                     & \textbf{In-Context RALM} & 0.267                 & 0.376                & 0.243             & 0.341             & 0.641               & 0.176           & 0.213         &0.322   \\
                                     & \textbf{IRCoT}           & 0.268                 & 0.376                & 0.216             & 0.314             & 0.613               & 0.188           & 0.226         &0.314   \\
                                     & \textbf{FLARE}           & 0.197                 & 0.276                & 0.246             & 0.341             & 0.609               & 0.183           & 0.214         &0.295   \\
                                     & \textbf{DRAGIN}          & 0.212                 & 0.302                & 0.272    & 0.378    & 0.662               & 0.189           & 0.226         & 0.320   \\
                                     & \textbf{ETC(Ours)}       & \textbf{0.352}        & \textbf{0.453}       & 0.272             & \textbf{0.487}             & \textbf{0.672}      & \textbf{0.286}  & \textbf{0.328}  & \textbf{0.420\textsubscript{+12.0\%}}\\\hline
\multirow{7}{*}{\textbf{Vicuna-13b}} & \textbf{w/o RAG}         & 0.146                 & 0.223                & 0.228             & 0.326             & 0.682               & 0.175           & 0.215          &0.285  \\
                                     & \textbf{Single RAG}      & 0.170                 & 0.256                & 0.254             & 0.353             & 0.686               & 0.217           & 0.256         &0.313   \\
                                     & \textbf{In-Context RALM} & 0.135                 & 0.213                & 0.187             & 0.304             & 0.645               & 0.099           & 0.129          &0.245  \\
                                     & \textbf{IRCoT}           & 0.188                 & 0.263                & 0.185             & 0.322             & 0.622               & 0.103           & 0.134          &0.260  \\
                                     & \textbf{FLARE}           & 0.157                 & 0.226                & 0.092             & 0.181             & 0.599               & 0.117           & 0.147           &0.217 \\
                                     & \textbf{DRAGIN}          & 0.252                 & 0.352                & 0.288             & 0.416             & 0.687               & \textbf{0.223}  & 0.265          &\underline{0.355}  \\
                                     & \textbf{ETC(Ours)}       & \textbf{0.282}        & \textbf{0.373}       & \textbf{0.347}    & \textbf{0.456}    & \textbf{0.693}      & 0.216           & \textbf{0.268}  &\textbf{0.376\textsubscript{\scriptsize{+5.9\%}}} \\ \bottomrule[1.5pt] 
\end{tabular}
\caption{The main results of various RAG methods. All results are obtained from publicly available papers or reproduced using open-source code. The best result of each dataset is in bold. To minimize randomness, we tested each model three times and reported the average performance.Additionally, we conducted t-tests to compare our results with previous results, confirming that our results are statistically significant with a $p$-value of less than 0.05.}
\label{tab:main}
\end{table*}

\subsection{Experimental Details}
To ensure a fair comparison with previous works, all experimental settings are the same as FLARE~\cite{jiang2023active} and DRAGIN~\cite{su2024dragin}. Specifically, we follow the setting of \citet{wang2022self} to generate both chain-of-thought (CoT) reasoning process as well as the final answer, we also use prompt templates from \citet{trivedi2022interleaving,jiang2023active,wei2022chain} tailored to each dataset. We use BM25 as the retriever due to its high efficiency and strong retrieval performance. To compute the entropy sequence, we remove stop words using the SpaCy library~\footnote{https://spacy.io/}. Wikipedia is used as the external knowledge corpus, from which we retrieve three augmented passages at each retrieval step. The backbone LLMs include LLaMa2-7b, LLaMa2-13b~\cite{touvron2023llama}, LLaMa3-8b~\cite{grattafiori2024llama} and Vicuna-13b-v1.5~\cite{chiang2023vicuna}, we report the results of LLaMa2-13B in the Appendix due to space limitations. 

\subsection{Comparison Models}

Since ETC is a training-free dynamic RAG method, it does not need any pre-training or fine-tuning process. In this paper, we compare ETC with the following advanced training-free RAG methods. 1) \textbf{w/o RAG} directly asks LLMs to generate answers without any retrieval operation; 2) \textbf{Single RAG} retrieves augmented information only once at the beginning of the generation based on the initial question; 3) \textbf{In-Context RALM}~\cite{ram2023context} triggers the retrieval module every $n$ tokens; 4) \textbf{IRCoT}~\cite{trivedi2022interleaving} activates the retrieval module every sentence; 5) \textbf{FLARE}~\cite{jiang2023active} conducts retrieve when a token's uncertainty below a threshold; 6) \textbf{DRAGIN}~\cite{su2024dragin} further considers both the importance and uncertainty of the generated token to determine the retrieval timing.

Besides the above previous works, we built several model variants for ablation study: 1) \textbf{ETC}$_{1st}$ indicates that we use the first difference and dynamic smoothing method to determine retrieval timing; 2) \textbf{ETC w/o smoothing} removes the dynamic smoothing method and relies solely on the second difference of the entropy sequence to trigger retrieval; 3) \textbf{ETC}$_{fixed}$ uses a fixed weight factor $w_t$ to smooth the second difference. 
\section{Main Results}
\subsection{Main Results}
Table 2 presents the main results, from which we draw three key findings. (1) Simple training-free RAG methods, such as In-Context RALM and IRCoT, which perform retrieval at fixed intervals (e.g., every $n$ tokens or each sentence), fail to consistently improve performance across tasks or models. This supports prior findings that indiscriminate or misaligned retrieval may degrade generation quality. (2) ETC consistently achieves the best overall performance across all evaluated settings. It outperforms both static and dynamic RAG baselines, achieving the highest average scores on each model: 0.344 on LLaMA2-7B, 0.420 on LLaMA3-8B, and 0.376 on Vicuna-13B, with relative improvements ranging from 5.9\% to 12.1\% over the strongest competing methods. These results validate that ETC’s trend-based retrieval mechanism enables more accurate and timely intervention compared to token-level confidence thresholds. (3) ETC further demonstrates strong adaptability across different task types and model scales. Notably, on LLaMA3-8B, the Single RAG baseline achieves competitive performance (0.375), surpassing several dynamic RAG methods such as DRAGIN and FLARE. This suggests that in high-capacity models, suboptimal retrieval timing can disrupt the generation process, leading to performance degradation. In contrast, ETC maintains robust improvements (0.420), indicating that its trend-aware strategy scales effectively with stronger LLMs.

Overall, ETC mitigates delayed retrieval by modeling confidence trends, thereby enabling more effective integration of external knowledge and improving generation performance.


\begin{figure}[H]
    \centering
    \includegraphics[width=1.0\linewidth]{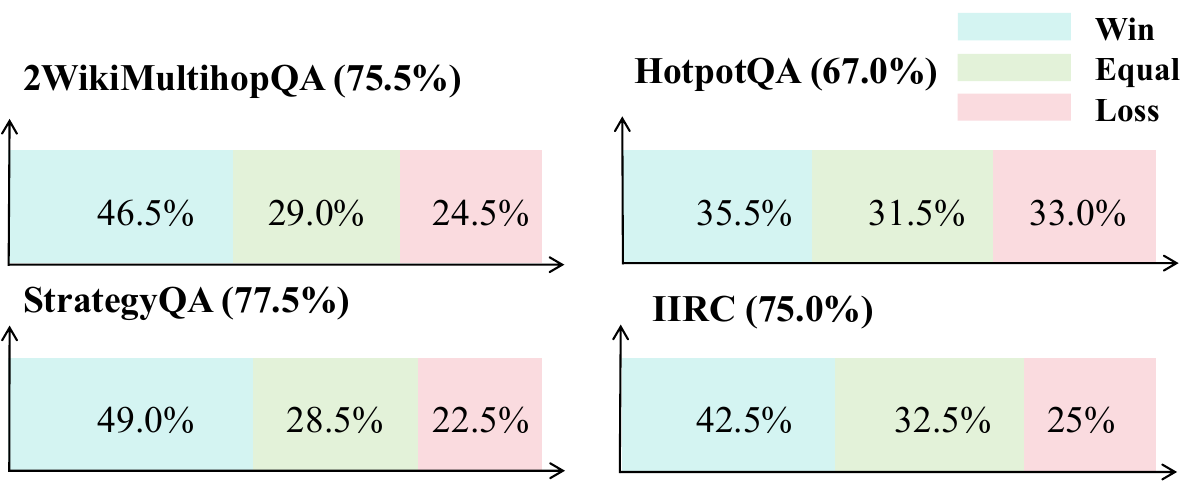}
    \caption{The win rate using GPT-4o as judge. The value in each bracket indicates the percentage of times ETC's answer quality is equal to or better than DRAGIN's on the corresponding dataset.}
    \label{pic.win}
\end{figure}

\subsection{Win Rate}

Several studies~\cite{li2023generative,yang2024ucfe,li2024leveraging} suggest that traditional metrics such as EM and F1 alone may not fully capture the performance of generative models, as these models can produce semantically correct answers that do not exactly match human-labeled references at the token level. To better assess answer quality, we employ GPT-4o as an evaluator. Specifically, we randomly sample 200 instances from each dataset and ask GPT-4o to judge which answer is more reasonable based on the given question and the ground-truth answer. The evaluation prompt instructs GPT-4o to assess responses based on accuracy, completeness, fluency, and relevance~\footnote{The full prompt is provided in Appendix A}. The results in Figure~\ref{pic.win} show that ETC consistently achieves performance comparable to or better than DRAGIN across most datasets. This confirms that ETC generates more reasonable answers than previous SoTA models, further demonstrating its effectiveness as a dynamic RAG system.


\subsection{Domain-Specific Dataset Evaluation}

RAG methods are particularly valuable for injecting external knowledge when LLMs lack domain-specific expertise, making them essential for solving domain-specific tasks. To assess their effectiveness in domain-specific scenarios, we evaluate RAG methods on two biomedical datasets: BioASQ and PubMedQA. As shown in Table~\ref{tab:domain}, ETC achieves substantial improvements over existing RAG models with fewer retrievals, demonstrating its effectiveness in domain-specific tasks. These findings suggest that ETC triggers retrieval at the appropriate moment, ensuring external knowledge is injected when needed, thereby reducing hallucinations and enhancing domain-specific understanding.

\begin{table}[!h]
\centering
\begin{tabular}{@{}c|cc|cc@{}}
\toprule[1.5pt]
\textbf{}           & \multicolumn{2}{c|}{\textbf{BioASQ}}          & \multicolumn{2}{c}{\textbf{PubmedQA}}     \\ \hline
\textbf{Method}     & \textbf{Acc.} & \textbf{Count.} & \textbf{Acc.} & \textbf{Count.} \\\midrule[1.5pt]
\textbf{Single RAG} & 0.478         & 1.0                       & 0.485         & 1.0                       \\
\textbf{IRCoT}               & 0.319         & 6.038                     & 0.397         & 6.433                     \\
\textbf{DRAGIN}     & 0.527         & 1.849                     & 0.510         & 1.898                     \\
\textbf{ETC}        & \textbf{0.689\textsubscript{\scriptsize{+30.7\%}}}         & 1.495                     & \textbf{0.545\textsubscript{\scriptsize{+6.9\%}}}         & 1.747                     \\ \bottomrule[1.5pt]
\end{tabular}
\caption{Accuracy (\textbf{Acc.}) and retrieval count (\textbf{Count.}) on domain-specific datasets. We use Vicuna-13b here since it achieves the best results in the main evaluation.}
\label{tab:domain}
\end{table}

\begin{table*}[ht]
\centering
\setlength{\tabcolsep}{1.2mm}
\begin{tabular}{c|cc|cc|c|cc|c}
\toprule[1.5pt]
                       & \multicolumn{2}{c|}{\textbf{2WikiMultihopQA}} & \multicolumn{2}{c|}{\textbf{HotpotQA}} & \textbf{StrategyQA} & \multicolumn{2}{c|}{\textbf{IIRC}} &                                       \\ \cline{2-8}
\multirow{-2}{*}{}     & \textbf{EM}           & \textbf{F1}          & \textbf{EM}       & \textbf{F1}       & \textbf{ACC}        & \textbf{EM}     & \textbf{F1}     & \multirow{-2}{*}{\textbf{Avg.Score}}  \\\midrule[1.5pt]
\textbf{ETC}$_{1st}$ & 0.271                 & 0.364               & 0.275             & 0.388             & 0.658               & 0.198          & 0.234          & 0.341                                 \\
\textbf{ETC}$_{fixed}$          & 0.271                 & 0.360               & 0.261             & 0.371             & 0.650                & 0.194          & 0.229          & 0.334                                 \\
\textbf{ETC}                    & 0.271                 & 0.360               & 0.288             & 0.401            & 0.650                & 0.199          & 0.240          &\textbf{0.344} \\
\textbf{ETC w/o smoothing}      & 0.269                 & 0.358               & 0.269             & 0.376            & 0.641               & 0.193         & 0.230          & 0.334  \\\bottomrule[1.5pt]                              
\end{tabular}
\caption{Ablation studies on various components in ETC with LLaMA2-7B-chat as backbone, other LLMs show similar results. We set the fixed weight in \textbf{ETC}$_{fixed}$ as 0.9 here.}
\label{tab:ablation}
\end{table*}

\section{Analysis}

\subsection{Ablation Study}
We conduct ablation studies to assess the effectiveness of using the second difference and the dynamic smoothing strategy. Table~\ref{tab:ablation} shows the results and we have the following observations: 1) Compared to DRAGIN, which triggers retrieval based on single-token confidence, and ETC$_{1st}$ which uses only the first difference of the entropy sequence, using the second difference yields significantly better average scores; 2) Removing the dynamic smoothing module or replacing it with a fixed weight factor (e.g., 0.9 in our experiment) results in consistent performance degradation across most datasets. The above results verify the effectiveness of our proposed components in handling various question answering scenarios.

\subsection{Retrieval Efficiency}
The average number of retrievals serves as a key metric for assessing the efficiency of dynamic RAG methods. As shown in Table~\ref{tab:count}, ETC consistently requires fewer retrievals than prior dynamic RAG baselines such as DRAGIN and FLARE. Moreover, removing the dynamic smoothing module or replacing it with a fixed weight leads to an increased number of retrievals. This validates that relying solely on entropy trends may result in redundant retrievals. The dynamic smoothing module effectively reduces unnecessary retrievals by mitigating the impact of outliers in the entropy sequence, thereby improving overall RAG efficiency\footnote{We report detailed retrieval counts for each dataset with each LLM in Appendix C.}. In addition, we compute average delayed length, and results show that DRAGIN retrieves on average 8.64 tokens later than ETC. This further confirms that ETC is not only more accurate in deciding whether to retrieve, but also significantly more timely.

\begin{table}[H]
\centering
\begin{tabular}{@{}c|c|c|c|c|c@{}}
\toprule[1.5pt]
                           & \multicolumn{1}{c|}{\textbf{WQA}} & \multicolumn{1}{c|}{\textbf{HQA}} & \multicolumn{1}{c|}{\textbf{SQA}} & \multicolumn{1}{c|}{\textbf{IIRC}} & \multicolumn{1}{c}{\textbf{Avg.R}} \\ \midrule[1.5pt]
 \textbf{FLARE}                     & 1.21                                        & 1.84                                 & 1.01                                   & 2.37                             & 1.59                                \\                          
\textbf{DRAGIN}                     & 2.67                                        & 3.23                                 & 4.39                                   & 2.96                             & 3.31                                \\
\textbf{ETC}$_{fixed}$ & 1.56                                        & 1.07                                  & 1.64                                    & 1.64                              & 1.47                                  \\
\textbf{ETC}                        & 1.43                                        & 0.88                                 & 1.37                                   & 1.48                             & \textbf{1.29}                        \\
\textbf{w/o smoothing}          & 1.47                                        & 0.92                                 & 1.47                                   & 1.52                              & 1.34                                 \\ \bottomrule[1.5pt]
\end{tabular}
\caption{The average retrieval count for each dataset, where \textbf{Avg.R} denotes the average retrieval count across datasets and LLMs. To conserve space, we abbreviate 2WikiMultihopQA, HotpotQA, and StrategyQA as WQA, HQA, and SQA, respectively.}
\label{tab:count}
\end{table}

\subsection{Selecting Optimal Retrieval Timing}\label{sec.delay} 

We further investigate whether ETC can mitigate the delayed retrieval and the redundant retrievals. To evaluate retrieval timeliness, we randomly select 100 samples from 2WikiMultihopQA and ask three experts to manually assess whether retrieval operations occurred at the appropriate moment. For the \textbf{delayed retrieval}, a retrieval operation is considered delayed if incorrect tokens are generated before retrieval occurs; otherwise, it is classified as timely. The results in Table~\ref{tab:timing} indicate that DRAGIN exhibits high delayed retrieval ratios, while ETC achieves the lowest delay ratio. This is because the second difference effectively captures the rate of confidence change, making it more sensitive to rapid fluctuations. This enables earlier retrieval interventions before low-confidence tokens are generated. As for \textbf{redundant retrieval}, a retrieval is considered redundant if the LLM could generate the correct answer even without retrieving external information. We can observe from Table~\ref{tab:timing} that ETC shows a significantly lower redundant retrieval ratio than DRAGIN and ETC w/o smoothing; this verifies the usefulness of the dynamic smoothing module in improving retrieval efficiency.  

\begin{table}[!h]
\centering
\begin{tabular}{@{}c|cc@{}}
\toprule[1.5pt]
                       & \textbf{\begin{tabular}[c]{@{}c@{}}Delayed \\ Retrieval Ratio\end{tabular}} & \textbf{\begin{tabular}[c]{@{}c@{}}Redundant \\ Retrieval Ratio\end{tabular}} \\ \midrule[1.5pt]
\textbf{DRAGIN}        &   0.33      &     0.95                                                        \\
\textbf{ETC}        &    \textbf{0.22}                                                         &    \textbf{0.79}                                                           \\
\textbf{w/o smoothing} & 0.22                                                      &    0.91                                                                     \\ \bottomrule[1.5pt]
\end{tabular}
\caption{The manually annotated delayed retrieval ratio and redundant retrieval ratio, and lower is better.}
\label{tab:timing}
\end{table}

\subsection{The Heat-map of Retrieval Timing and The Entropy Distribution}

In this subsection, we present a heat-map illustrating the word positions where ETC and DRAGIN trigger their first retrieval. Additionally, we provide the average entropy values for each word position. The visualization shows that ETC typically triggers retrieval earlier than DRAGIN, which aligns with the trend of gradually increasing word entropy. In contrast, DRAGIN often delays retrieval until encountering highly uncertain words or even later.


\begin{figure}[H]
    \centering
    \includegraphics[width=1.0\linewidth]{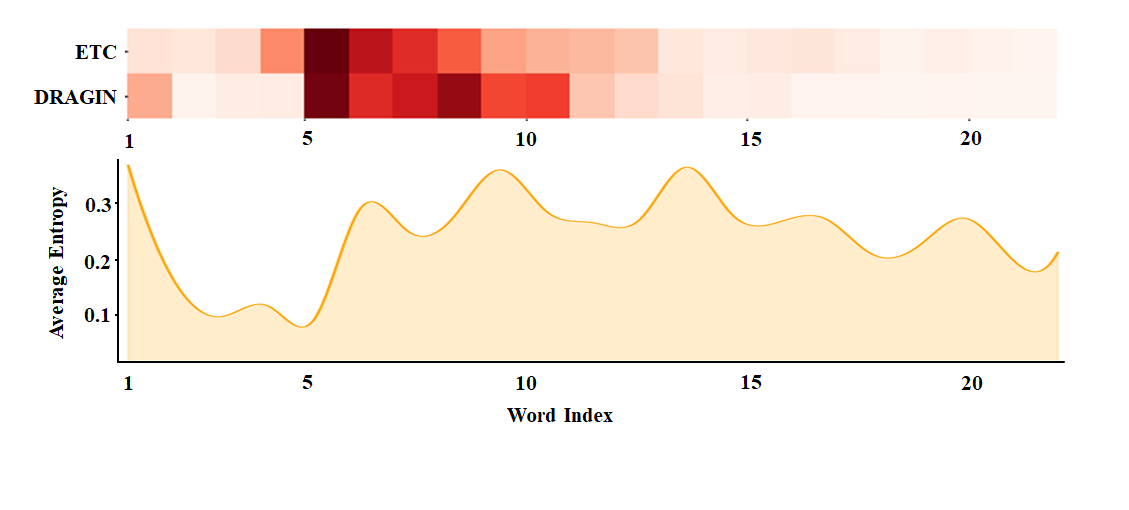}
    \caption{The heat-map of retrieval timing and the entropy distribution.}
    \label{pic.heat}
\end{figure}

\begin{figure*}[!h]
    \centering
    \includegraphics[width=1.0\linewidth]{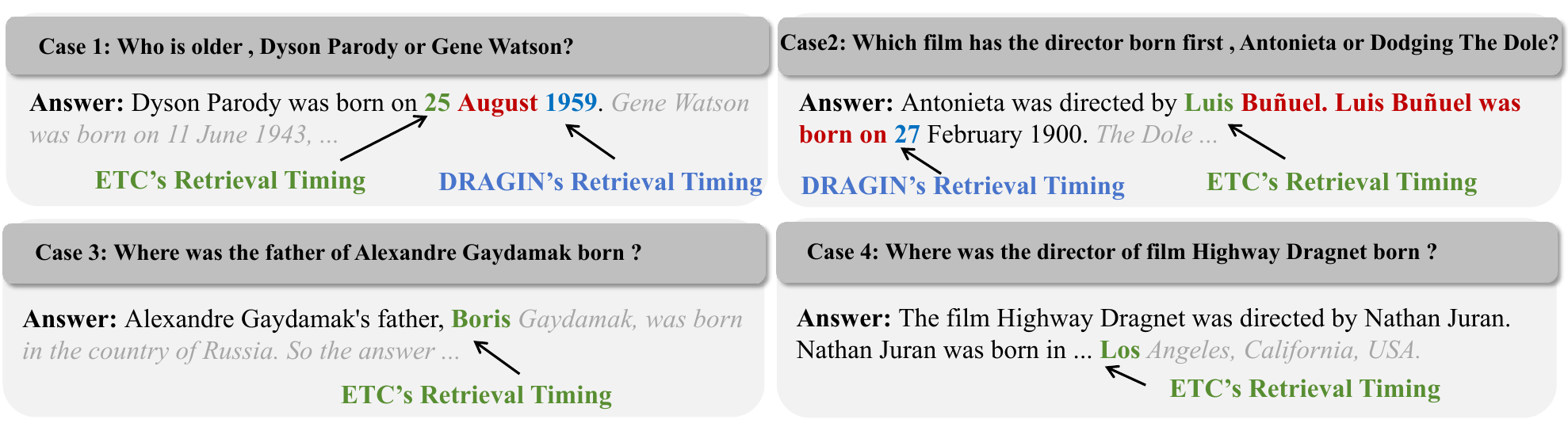}
    \caption{Illustrative cases of delayed retrieval. The first two cases demonstrate delayed retrieval, where green tokens indicate ETC's retrieval timing, blue tokens represent DRAGIN's retrieval timing, and red tokens highlight incorrectly generated tokens caused by delayed retrieval. The last two cases illustrate missing retrieval, which is a special case of delayed retrieval.}
    \label{pic.case}
\end{figure*}

Interestingly, while DRAGIN appears redder at the first position in the heat-map, which may suggest early retrieval, a closer analysis reveals that most of these early retrievals are redundant and ineffective. Specifically, in 54 samples where DRAGIN retrieves at the first token, only 10 result in improved answers, while the remaining 44 are equal to or worse than those without retrieval, yielding a redundancy rate of 81.5\%. This observation is consistent with the high redundant retrieval ratio reported in Table~\ref{tab:timing}.

In contrast, ETC not only triggers retrieval at earlier average positions but also achieves more effective retrieval by leveraging meaningful uncertainty trends. These findings confirm ETC’s advantage in both timing precision and retrieval effectiveness. 


\subsection{Case Study}
In addition to the quantitative analysis, we present several intuitive cases from 2WikiMultihopQA to illustrate how different dynamic RAG methods behave during the generation process, as shown in Figure~\ref{pic.case}. 

The first two cases in Figure~\ref{pic.case} show that DRAGIN generates multiple incorrect tokens before triggering retrieval, exhibiting the delayed retrieval issue. This occurs because DRAGIN determines retrieval timing based solely on single-token confidence, which may fail to trigger retrieval early enough when external knowledge is required. In contrast, ETC leverages entropy change trends to intervene at the right moment, leading to more coherent and accurate generation with timely access to external knowledge.

We also identify a special form of delayed retrieval, referred to as missing retrieval, where the dynamic RAG system fails to trigger retrieval at all during the generation process. The last two cases in Figure~\ref{pic.case} show that DRAGIN fails to detect the appropriate retrieval timing, whereas ETC intervenes effectively, resulting in the correct answer.


\section{Related Work}

Retrieval augmented generation is an efficient and effective approach to help LLMs obtaining necessary external knowledge\cite{fan2024survey}. Existing works mainly focusing on training-based RAG system~\cite{Yoran2023MakingRL,Luo2024LandmarkEA,fang2024enhancingnr,xu2024unsupervisedir} and training-free RAG system~\cite{Izacard2020LeveragingPR,Wang2023SelfKnowledgeGR,jiang2023active,su2024dragin}. This paper focuses on the later one since it is more lightweight and efficient in practical scenarios.

In the era of LLM, early research mainly explores designing more suitable prompts for high-quality retrieved text~\cite{Shi2023REPLUGRB,Wang2023Query2docQE,Yu2023ImprovingLM},these methods typically conduct retrieval operations only once at the start of the generation process. Lately, people found that not all retrieval operations are beneficial for LLM’s generation, improper or redundant augmented information may cause negative influence on the performance~\cite{Wang2023SelfKnowledgeGR,ni2024whendl,Su2024BRIGHTAR}. Based on the above observation, more research explore to active the retrieval operation when LLM needed, named as dynamic RAG. \citet{borgeaud2022improving,trivedi2022interleaving,ram2023context} proposed to retrieve every $n$ tokens or every sentence, making LLM receive new knowledge during generation process. While \citet{jiang2023active,Wang2024SelfDCWT,Tao2024WhenTT} propose to determine the retrieval timing based on the prediction confidence of the generated token or the internal states. \citet{su2024dragin} further considers the importance of each token to find more reasonable retrieval timing.

\section{Conclusion}

In this paper, we introduced Entropy-Trend Constraint (ETC), a training-free method for selecting optimal retrieval timing in dynamic retrieval-augmented generation. Unlike prior approaches that rely on a single token's confidence, ETC models entropy trends over time to detect rising uncertainty and trigger retrieval more effectively. Extensive experiments on six QA datasets across multiple LLMs demonstrate that ETC consistently outperforms strong baselines while reducing retrieval frequency. Beyond performance improvements, our findings shed light on the role of temporal uncertainty modeling in retrieval-aware generation, offering practical guidance for designing future dynamic RAG systems.

\section*{Acknowledgements}
This work was supported by the National Natural Science Foundation of China (Grant No. 62276089 and Grant No. 62506115), the Natural Science Foundation of Tianjin (Grant No. 24JCJQJC00200 and Grant No. 24JCQNJC01230), the Natural Science Foundation of Hebei Province (Grant No. F2024202064 and Grant No. F2025202020), the Science Research Project of Hebei Education Department (Grant No. BJ2025004), the Ministry of Human Resources and Social Security of China (Grant No. RSTH-2023-135-1), and the Science and Technology Program of Hebei Province (Grant No. 24464401D).

\bibliography{aaai2026}

\clearpage
\section*{Appendix}
\subsection*{Appendix A: Prompt Used in Win Rate Evaluation}\label{app:prompt}

For a more comprehensive evaluation of answer's quality using GPT-4o, we design the evaluation prompt containing the following aspects: accuracy, completeness, fluency, and relevance. We randomly select 100 instances from each dataset for evaluation. Specially, we ask GPT-4o to compare the overall quality of ETC's answer and DRAGIN's answer using the following prompt:

\begin{figure}[!h]
    \centering
    \includegraphics[width=1.0\linewidth]{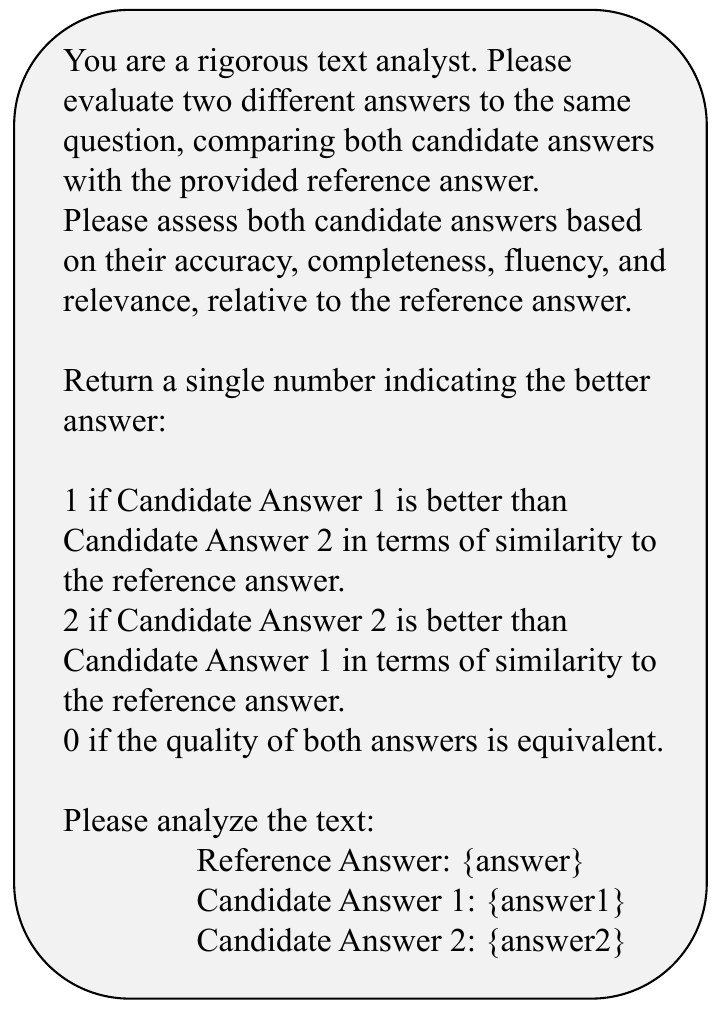}
    \caption{The prompt used to evaluate the answer quality in our paper.}
    \label{pic:prompt}
\end{figure}

\subsection*{Appendix B: Hyper-parameter Used in ETC}\label{app:parma}

We list the parameters used in our method in Table~\ref{tab:app.param}, all chosen parameters are determined on the validation  set of each dataset. We also conducted several ablation study to test the sensitivity of $\alpha$, the results shows that we could obtain similar or even better performance around the selected $\alpha$.

\begin{table*}[]
\centering
\begin{tabular}{c|c|cccc}
\toprule[1.5pt]
                                          & \textbf{Hyper-parameters} & \textbf{2WikiMultihopQA} & \textbf{HotpotQA} & \textbf{StrategyQA} & \textbf{IIRC} \\\midrule[1.5pt]
\multirow{2}{*}{\textbf{Llama-2-7b}} &       $\alpha$                   & 1.0                      & 1.3               & 0.75                & 1.0           \\
                                          & top $n$ tokens             & 25                       & 35                & 35                  & 35            \\\hline
\multirow{2}{*}{\textbf{Llama-3-8b}} &       $\alpha$                   & 1.0                      & 1.3               & 0.75                & 1.0           \\
                                          & top $n$ tokens             & 25                       & 35                & 35                  & 35            \\\hline
\multirow{2}{*}{\textbf{Vicuna-13b}} &        $\alpha$                  & 1.2                      & 1.2               & 1.5                 & 1.0           \\
                                          & top $n$ tokens             & 25                       & 35                & 35                  & 35  \\\bottomrule[1.5pt]                              
\end{tabular}
\caption{The parameters used in ETC. We choose the $\alpha$ on the validation set of each dataset, and top $n$ tokens is the same as used in DRAGIN paper.}
\label{tab:app.param}
\end{table*}

\subsection*{Appendix C: The Results with LLaMa2-13b}\label{app:13b}

We test our method based on LLaMa2-13b and the results draw similar conclusions.

\begin{table*}[!h]
\centering
\setlength{\tabcolsep}{1.5mm}
\begin{tabular}{c|c|cc|cc|c|cc}
\toprule[1.5pt]
                                     &                          & \multicolumn{2}{c|}{\textbf{2WikiMultihopQA}} & \multicolumn{2}{c|}{\textbf{HotpotQA}} & \textbf{StrategyQA} & \multicolumn{2}{c}{\textbf{IIRC}} \\ \hline
\textbf{LLM}                         & \textbf{RAG Method}      & \textbf{EM}           & \textbf{F1}          & \textbf{EM}       & \textbf{F1}       & \textbf{Accuracy}   & \textbf{EM}     & \textbf{F1}     \\\midrule[1.5pt]
\multirow{7}{*}{\textbf{Llama2-13b}} & \textbf{w/o RAG}         & 0.187                 & 0.272                & 0.223             & 0.310             & 0.650               & 0.168           & 0.204           \\
                                     & \textbf{Single RAG}      & 0.245                 & 0.336                & 0.263             & 0.371             & 0.654               & 0.196           & 0.230           \\
                                     & \textbf{In-Context RALM} & 0.217                 & 0.305                & 0.177             & 0.268             & 0.648               & 0.155           & 0.188           \\
                                     & \textbf{IRCoT}           & 0.270                 & 0.361                & 0.267             & 0.372             & 0.655               & 0.171           & 0.206           \\
                                     & \textbf{FLARE}           & 0.224                 & 0.308                & 0.180             & 0.276             & 0.655               & 0.138           & 0.167           \\
                                     & \textbf{DRAGIN}          & 0.304                 & 0.393                & \textbf{0.314}    & \textbf{0.424}    & 0.689               & 0.185           & 0.222           \\
                                     & \textbf{ETC(Ours)}       & \textbf{0.311}        & \textbf{0.405}       & 0.275             & 0.374             & \textbf{0.693}      & \textbf{0.215}  & \textbf{0.250}  \\\bottomrule[1.5pt] 
\end{tabular}
\caption{The main results of various RAG methods with LLaMa2-13B as backbone.}
\label{tab:main}
\end{table*}

\subsection*{Appendix D: The Sensitivity Analysis of $\alpha$}

We also test our method with different $\alpha$.

\begin{table*}[]
\centering
\setlength{\tabcolsep}{1.0mm}
\begin{tabular}{l|c|cccccccccccc}
\hline
\multirow{9}{*}{\textbf{Llama2-7b}} & $\alpha$                                 & \multicolumn{2}{c}{\textit{2.2}} & \multicolumn{2}{c}{\textit{1.9}} & \multicolumn{2}{c}{\textit{1.6}} & \multicolumn{2}{c}{\textit{1.3}} & \multicolumn{2}{c}{\textit{1.0}} & \multicolumn{2}{c}{\textit{0.7}} \\ \cline{2-14} 
                                    & \multirow{2}{*}{\textbf{2WikimultihopQA}} & \textbf{EM}     & \textbf{F1}    & \textbf{EM}     & \textbf{F1}    & \textbf{EM}     & \textbf{F1}    & \textbf{EM}     & \textbf{F1}    & \textbf{EM}     & \textbf{F1}    & \textbf{EM}     & \textbf{F1}    \\
                                    &                                           & 0.252           & 0.344          & 0.259           & 0.347          & 0.265           & 0.352          & 0.269           & 0.352          & 0.271           & 0.360          & 0.268           & 0.355          \\ \cline{2-14} 
                                    & \multirow{2}{*}{\textbf{HotpotQA}}        & \textbf{EM}     & \textbf{F1}    & \textbf{EM}     & \textbf{F1}    & \textbf{EM}     & \textbf{F1}    & \textbf{EM}     & \textbf{F1}    & \textbf{EM}     & \textbf{F1}    & \textbf{EM}     & \textbf{F1}    \\
                                    &                                           & 0.258           & 0.363          & 0.266           & 0.367          & 0.274           & 0.387          & 0.288           & 0.401          & 0.287           & 0.400          & 0.287           & 0.400          \\ \cline{2-14} 
                                    & \multirow{2}{*}{\textbf{StrategyQA}}      & \multicolumn{2}{c}{\textbf{ACC}} & \multicolumn{2}{c}{\textbf{ACC}} & \multicolumn{2}{c}{\textbf{ACC}} & \multicolumn{2}{c}{\textbf{ACC}} & \multicolumn{2}{c}{\textbf{ACC}} & \multicolumn{2}{c}{\textbf{ACC}} \\
                                    &                                           & \multicolumn{2}{c}{0.650}        & \multicolumn{2}{c}{0.648}        & \multicolumn{2}{c}{0.647}        & \multicolumn{2}{c}{0.647}        & \multicolumn{2}{c}{0.645}        & \multicolumn{2}{c}{0.642}        \\ \cline{2-14} 
                                    & \multirow{2}{*}{\textbf{IIRC}}            & \textbf{EM}     & \textbf{F1}    & \textbf{EM}     & \textbf{F1}    & \textbf{EM}     & \textbf{F1}    & \textbf{EM}     & \textbf{F1}    & \textbf{EM}     & \textbf{F1}    & \textbf{EM}     & \textbf{F1}    \\
                                    &                                           & 0.166           & 0.198          & 0.178           & 0.211          & 0.192           & 0.226          & 0.190           & 0.225          & 0.196           & 0.238          & 0.199           & 0.240          \\ \hline
\multirow{9}{*}{\textbf{Llama3-8b}} & $\alpha$                                  & \multicolumn{2}{c}{\textit{2.2}} & \multicolumn{2}{c}{\textit{1.9}} & \multicolumn{2}{c}{\textit{1.6}} & \multicolumn{2}{c}{\textit{1.3}} & \multicolumn{2}{c}{\textit{1.0}} & \multicolumn{2}{c}{\textit{0.7}} \\ \cline{2-14} 
                                    & \multirow{2}{*}{\textbf{2WikimultihopQA}} & \textbf{EM}     & \textbf{F1}    & \textbf{EM}     & \textbf{F1}    & \textbf{EM}     & \textbf{F1}    & \textbf{EM}     & \textbf{F1}    & \textbf{EM}     & \textbf{F1}    & \textbf{EM}     & \textbf{F1}    \\
                                    &                                           & 0.332           & 0.435          & 0.336           & 0.437          & 0.345           & 0.448          & 0.352           & 0.453          & 0.351           & 0.452          & 0.345           & 0.448          \\ \cline{2-14} 
                                    & \multirow{2}{*}{\textbf{HotpotQA}}        & \textbf{EM}     & \textbf{F1}    & \textbf{EM}     & \textbf{F1}    & \textbf{EM}     & \textbf{F1}    & \textbf{EM}     & \textbf{F1}    & \textbf{EM}     & \textbf{F1}    & \textbf{EM}     & \textbf{F1}    \\
                                    &                                           & 0.259           & 0.341          & 0.265           & 0.376          & 0.269           & 0.386          & 0.270           & 0.451          & 0.272           & 0.487          & 0.268           & 0.471          \\ \cline{2-14} 
                                    & \multirow{2}{*}{\textbf{StrategyQA}}      & \multicolumn{2}{c}{\textbf{ACC}} & \multicolumn{2}{c}{\textbf{ACC}} & \multicolumn{2}{c}{\textbf{ACC}} & \multicolumn{2}{c}{\textbf{ACC}} & \multicolumn{2}{c}{\textbf{ACC}} & \multicolumn{2}{c}{\textbf{ACC}} \\
                                    &                                           & \multicolumn{2}{c}{0.665}        & \multicolumn{2}{c}{0.664}        & \multicolumn{2}{c}{0.667}        & \multicolumn{2}{c}{0.669}        & \multicolumn{2}{c}{0.672}        & \multicolumn{2}{c}{0.665}        \\ \cline{2-14} 
                                    & \multirow{2}{*}{\textbf{IIRC}}            & EM              & F1             & EM              & F1             & EM              & F1             & EM              & F1             & EM              & F1             & EM              & F1             \\
                                    &                                           & 0.248           & 0.291          & 0.253           & 0.297          & 0.265           & 0.307          & 0.279           & 0.322          & 0.286           & 0.325          & 0.286           & 0.328          \\ \hline
\end{tabular}
\caption{The sensitivity analysis of $\alpha$.}
\end{table*}

\end{document}